\documentclass{ecai2012}
\usepackage{times}
\usepackage{amsmath}
\usepackage{graphicx}
\usepackage{xspace}
\usepackage{latexsym}
\newcommand{\mymin}{\mbox{\rm min}}
\newcommand{\mytext}[1]{\mbox{\rm #1}}

\newcommand{\winner}[2]{\mbox{$#1 + #2$}}
\newcommand{\winners}[3]{\mbox{$#1 + #2 + #3$}}
\newcommand{\topk}{\mbox{$+$}}

\newtheorem{proposition}{Proposition}
\newcommand{\myproof}{\vspace{-3mm}\noindent {\bf Proof:\ \ }}
\newcommand{\myqed}{\mbox{$\Box$}}

\newcommand{\plurshort}{\mbox{\emph{Plur}}\xspace}
\newcommand{\bordashort}{\mbox{\emph{Greedy}}\xspace}
\newcommand{\bordaplurshort}{\mbox{\emph{AdaptGreedy}}\xspace}

\newcommand\lirong[1]{}

\newtheorem{dfn}{Definition}

\begin{document}

\title{Combining Voting Rules Together}

\author{Nina Narodytska\institute{NICTA and UNSW, Sydney, Australia, email: ninan@cse.unsw.edu.au} \and Toby Walsh\institute{NICTA and UNSW, Sydney, Australia, email: toby.walsh@nicta.com.au} \and Lirong Xia\institute{ Harvard University, Cambridge, MA, USA, email: lxia@seas.harvard.edu} }

\maketitle
\bibliographystyle{ecai2012}

\begin{abstract}
We propose a simple method for combining together
voting rules that performs a run-off between
the different winners of each voting rule.
We prove that this combinator has several
good properties. For instance, even if just
one of the base voting rules has a desirable
property like Condorcet consistency,
the combination inherits this property. In addition, we prove
that combining voting rules together in this
way can make finding a manipulation more
computationally difficult. Finally,
we study the impact of this combinator
on approximation methods that
find close to optimal manipulations.
\end{abstract}

\section{INTRODUCTION}

An attractive idea in the Zeitgeist of
contemporary culture is ``The Wisdom of Crowds''
\cite{wisdomcrowds}. This is the idea
that, by bringing together diversity
and independence of opinions,
groups can be better at making decisions than
the individuals that make up the group.
For example, in 1907 Galton observed the
wisdom of the crowd at guessing the weight
of an ox in the West of England Fat Stock and
Poultry Exhibition. The median of the 787 estimates
was 1207 lb, within 1\% of the correct weight of
1198 lb.
%
%
We can view different voting rules as having
different opinions on the ``best'' outcome
to an election. We argue here
that it may pay to combine these different opinions together.
We provide both theoretical and experimental
evidence for this thesis. On the theoretical
side, we argue that a combination of voting rules
can inherit a desirable property like
Condorcet consistency when only one of the base
voting rules is itself Condorcet consistent.
We also prove that combining
voting rules together can make strategic voting
more computationally difficult.
On the experimental
side, we study the impact of combining
voting rules on the performance of approximation methods
for constructing manipulations.

\subsection{RELATED WORK}\lirong{Moved to here.}
Different ways of combining voting rules to make manipulation computationally hard have been investigated recently.
Conitzer and Sandholm
\cite{csijcai03} studied the impact on the
computational complexity of manipulation
of adding an initial round of the Cup rule to a voting
rule.
This initial round eliminates half the candidates and
makes manipulation NP-hard to compute
for several voting rule including plurality and Borda.
Elkind and Lipmaa
\cite{elisaac05} extended this idea to
a general technique for combining two voting rules.
The first voting rule is run for some number
of rounds to eliminate some of the candidates,
before the second voting rule is applied to
the candidates that remain. They proved that
many such combinations of voting rules are NP-hard
to manipulate.
Note that theirs is a sequential combinator,
in which the two rules are run in sequence,
whilst ours (as we will see soon) is a parallel combinator,
in which the two rules are run in parallel.
More recently, Walsh and Xia
\cite{wxaamas12} showed that
using a lottery to eliminate some of the voters
(instead of some of the candidates) is another
mechanism to make manipulation intractable to compute.

\section{VOTING RULES}

Voting is a general mechanism to combine together
the preferences of agents.
Many different voting rules have been proposed
over the years, providing
different opinions as to the ``best''
outcome of an election. We formalise voting as follows.
A {\em profile} is a sequence of $n$ total orders over $m$ candidates.
A {\em voting rule} is a function mapping a profile
onto one candidate, the {\em winner}.
Let $N_P(i,j)$ be
the number of voters preferring $i$ to $j$ in $P$.
Where $P$ is obvious from the context, we write $N(i,j)$.
Let $beats(i,j)$ be 1 iff $N(i,j) > \frac{n}{2}$
and 0 otherwise.
We consider some of the most common voting rules.

\noindent{\bf $\bullet$ Positional scoring rules:} Given a {\em scoring vector}
$(w_1,\ldots,w_m)$
of weights, the $i$th candidate in a vote
scores $w_i$, and the winner is the candidate with
highest total score over all the votes.
The {\bf plurality} rule has
the weight vector $(1,0,\ldots,0)$,
the {\bf veto} rule has
the vector $(1,1,\ldots,1,0)$,
the {\bf k-approval} rule has
the vector $(1,\ldots,1,0\ldots,0)$
containing $k$ 1s,
and
the {\bf Borda} rule has the vector $(m-1,m-2,\ldots,0)$.

\noindent{\bf $\bullet$ Cup:} The winner is the result of a series of pairwise
majority elections between candidates.
Given the \emph{agenda}, a binary tree
in which the roots are labelled with candidates, we label
the parent of two nodes by
the winner of the pairwise majority election between
the two children. The winner is the label of the
root.

\noindent{\bf $\bullet$ Copeland:} The candidate
with the highest Copeland score wins. The
Copeland score of candidate $i$
is $\sum_{j \neq i} beats(i,j)$.\lirong{I changed it to a more standard definition.}
The Copeland winner is the candidate that wins the
most pairwise elections.

\noindent{\bf $\bullet$ Maximin:} The maximin score of candidate
$i$ is $\mymin_{j \neq i} N(i,j)$. The candidate with the
highest maximin score wins.

\noindent{\bf $\bullet$ Single Transferable Vote (STV):} This rule
requires up to $m-1$ rounds. In each round,
the candidate with the least number of voters ranking
them first is eliminated until one of the remaining candidates
has a majority.

\noindent{\bf $\bullet$ Bucklin (simplified):} The Bucklin score of a candidate is the smallest $k$ such that the $k$-approval  score of the candidate is strictly larger than $n/2$. The candidate with the smallest Bucklin score wins.\lirong{Definition is changed.}


Note that in some cases, there can be
multiple winning candidates (e.g. multiple
candidates with the highest
Borda score). We therefore may
also need a tie-breaking mechanism.
All above voting rules can be extended to choose a winner for profiles with weights. In this paper, we study the {\em manipulation} problem (with weighted votes), defined as follows.
\begin{dfn}\lirong{I added a formal definition of the manipulation problem because some reviewer must ask for it...}
In a \emph{manipulation} problem, we are given an instance $(r,P^{NM},\vec
w^{NM},c,k,\vec w^M)$, where $r$ is a voting rule, $P^{NM}$ is the
non-manipulators' profile, $\vec w^{NM}$ represents the weights of
$P^{NM}$, $c$ is the alternative preferred by the manipulators, $k$
is the number of manipulators, and $\vec w^M=(w_1,\ldots,w_k)$
represents the weights of the manipulators. We are asked whether
there exists a profile $P^{M}$ of indivisible votes for the
manipulators such that $ c\in r((P^{NM},P^{M}), (\vec w^{NM},\vec
w^M))$.
\end{dfn}
When all weights equal to $1$, the problem is called manipulation with unweighted votes. In this paper, we assume that the manipulators controls the tie-breaking mechanism, that is, all ties are broken in favor of $c$.

A large number of normative properties
that voting rules might possess have been put forwards
including the following.

\noindent{\bf $\bullet$ Unanimity:}
If a candidate is ranked in the top place by all voters,  then this candidate wins.

\noindent{\bf $\bullet$ Monotonicity:}
If we move the winner up a voter's preference
order, while keeping preferences unchanged,  then the winner should not change.

\noindent{\bf $\bullet$ Consistency:}
If two sets of votes select the same
winner then the union of these two
sets should also select the same winner.

\noindent{\bf $\bullet$ Majority criterion:}
If the majority of voters rank a same candidate at the top, then
this candidate wins.

\noindent{\bf $\bullet$ Condorcet consistency:}
If a {\em Condorcet winner} exists (a candidate
who beats all others in pairwise elections) then
this candidate wins.

\noindent{\bf $\bullet$ Condorcet loser criterion:}
If a {\em Condorcet loser} exists (a candidate
who is beaten by all others in pairwise elections) then
this candidate does not win.

Such properties can be used to compare voting
rules. For example, whilst
STV satisfies the majority criterion,
Borda does not. On the other hand,
Borda is monotonic but STV is not.

\section{VOTING RULE COMBINATOR}

We consider a simple combinator, written $\topk$, for
combining together two or more voting rules.
This combinator collects together the set of winners
from the different rules.
If all rules agree, this is the overall winner.
Otherwise we recursively
call the combination of voting
rules on this restricted set of winning candidates.
If the recursion
does not eliminate any candidates,
we call some tie-breaking mechanism on
the remaining candidates.
For example, $\winner{plurality}{veto}$
collects together the plurality and veto winners of
an election. If they are the same candidate, then
this is the winner. Otherwise, there is a runoff
in which we call
$\winner{plurality}{veto}$ on the plurality and veto winners.
As both plurality and veto on two candidates compute
the majority winner, the overall winner of
$\winner{plurality}{veto}$ is
the winner of a majority election between
the plurality and veto winners.

This combinator has some simple algebraic properties.
For example, it is idempotent and commutative.
That is, $\winner{X}{X}=X$ and
$\winner{X}{Y}=\winner{Y}{X}$.
It has 
other more complex algebraic
properties. For example,
$\winner{(\winner{X}{Y})}{X}=\winner{X}{Y}$.
In addition, many normative properties are inherited from the
base voting rules.
Interestingly,
it is sometimes enough for just one of the base
voting rules to have a normative property
for the composition to have the same property.



\begin{proposition}
For unanimity, the majority criterion, Condorcet consistency,
and the Condorcet loser criterion,
if one of $X_1$ to $X_k$ and the tie-breaking
mechanism satisfy the property,
then $\winners{X_1}{\ldots}{X_k}$ also satisfy the same property.
\lirong{I commented out a seemingly duplicated sentence.}
\end{proposition}

On the other hand, there are some properties
which can be lost by the introduction of a run-off.

\begin{proposition}[Monotonicity]
$plurality$ and $Borda$ are both monotonic but
$\winner{plurality}{Borda}$ is not.
\end{proposition}
\myproof
Suppose we have 6 votes for
$b\succ c\succ a$,\lirong{There was a typo ($\prec$ should be $\succ$), which is corrected now.}
4 votes for $c \succ a \succ b$,
and 3 votes for  $a \succ b \succ c$ and 3 votes for
$a \succ c \succ b$.
Tie-breaking for both Borda and plurality is $c \succ a \succ b$.
Now $c$ is the Borda winner and $a$ is the plurality winner.
By tie-breaking, $c$ wins the run-off.
However, if we modify one
vote for $a \succ c \succ b$ to $c \succ a \succ b$,
then $b$ becomes the plurality winner
and wins the run-off.
Hence, $\winner{plurality}{Borda}$ is not
monotonic.
\myqed

We give a stronger 
result for
consistency. Scoring rules are consistent, but
the combination of any two different scoring rules
is not. By ``different rules'' we mean that there exists a profile for which these two rules select different winners. If two scoring rules are different, then their scoring vectors must be different. We note that the reverse is not true.

\begin{proposition}[Consistency]
Let $X$ and $Y$ be any two different
scoring rules, then
$\winner{X}{Y}$ is not consistent.
\end{proposition}
\myproof
Let $s(P,a)$ and $r(P,a)$ be
the score given to candidate $a$
by $X$ and $Y$ in profile $P$ respectively.
Since $X$ and $Y$ are different, there exists
$P_1$ over $a_1$ to $a_m$ such that $X$ on $P_1$
selects $a_1$ and $Y$ on $P_1$ selects $a_2$.
Then $s(P_1,a_1) > s(P_1,a_2)$ but
$r(P_1,a_1) < r(P_1,a_2)$.
WLOG suppose $a_1$ beats $a_2$ in pairwise
elections in $P_1$ and tie breaking elects
$a_1$ in favour of $a_2$ when they have the
same top score.
Let $P_2$ consist of $m$ votes $V_1$ to $V_m$
where for $i<m$, $V_i$ ranks $a_2$ in $i$th
place and $a_1$ in $i+1$th place, and
$V_m$ ranks $a_1$ in 1st place and $a_2$ in last
place.
Then $s(P_2,a_1) = s(P_2,a_2)$ and
$r(P_2,a_1) = r(P_2,a_2)$.
Let $k$ be such that $k(r(P_1,a_2)-r(P_1,a_1)) > r(V_m,a_1)-r(V_m,a_2)$,
and $P_3$ be the following profile of cyclic
permutations:
$a_1 \succ a_2 \succ a_3 \succ \ldots \succ a_m$,
$a_1 \succ a_2 \succ a_4 \succ \ldots \succ a_3$,
$\ldots$,
$a_1 \succ a_2 \succ a_m \succ \ldots \succ a_{m-1}$,
$a_2 \succ a_1 \succ a_3 \succ \ldots \succ a_m$,
$a_2 \succ a_1 \succ a_4 \succ \ldots \succ a_3$,
$\ldots$,
$a_2 \succ a_1 \succ a_m \succ \ldots \succ a_{m-1}$.
Let $P_4$ be $k$ copies of $P_2$,
and $P_5$ be $km$ copies of $P_1$,
$V_m$ and $km|P_1|$ copies of $P_3$.
Now $\winner{X}{Y}$ on $P_4$ or $P_5$ selects $a_1$ as winner.
But $\winner{X}{Y}$ on $P_4 \cup P_5$ selects $a_2$.
\myqed

It follows immediately that $\winner{plurality}{Borda}$
is not consistent.

\vspace{-2mm}
\section{STRATEGIC VOTING}
\vspace{-1mm}
Combining voting rules together
can hinder strategic
voting. One appealing escape from the Gibbard-Satterthwaite
theorem was proposed
by Bartholdi, Tovey and Trick 
\cite{bartholditoveytrick}.
Perhaps it is computationally so difficult to find a successful
manipulation that voters have little
option but to report their true preferences?
As is common in the literature, we consider two
different settings: unweighted votes where
the number of candidates is large and we have
just one or two manipulators, and weighted votes where
the number of candidates is small but we have
a 
coalition of manipulators.
Whilst unweighted votes are perhaps more common
in practice, the weighted case informs
us about the unweighted case when we have probabilistic
information about the votes \cite{csljacm07}.
Since there are many possible combinations of common
voting rules, we give a few illustrative results
covering some of the more interesting cases.
With unweighted votes, we prove that
computational resistance to manipulation is
typically inherited from the base rules.
With weighted votes, our results are stronger.
We prove that there are many combinations
of voting rules where the base rules are
polynomial to manipulate but their
combination is NP-hard. Combining voting
rules thus offers another mechanism to make manipulation
more computationally difficult.

\vspace{-2mm}
\subsection*{A FIRST OBSERVATION}\lirong{The results are not really general so I changed the caption...}
\vspace{-1mm}
It seems natural that the combination of voting
rules inherits the computational complexity of manipulating
the base rules. However, there is not a simple
connection between the computational complexity of the bases rules and their combination.
In this section, we show two examples of artificial voting rules to illustrate this discrepancy. In the first example, manipulation for the base rules are NP-hard, but manipulation for their combination can be computed in polynomial-time; in the second example, manipulation for the base rules are in P, but manipulation for their combination is NP-hard to compute.\lirong{Slightly changed the wording.}


\begin{proposition}
There exist voting rules $X$ and $Y$
for which computing a manipulation
is NP-hard but computing a manipulation
of $\winner{X}{Y}$
is polynomial.
\end{proposition}
\myproof
We give a reduction from 1 in 3-SAT on
positive clauses. Boolean variables
$1$ to $n$ are represented by the candidates
$1$ to $n$. We also have two additional
candidates $0$ and $-1$. Any vote
with $0$ in first place represents
a clause. The first three candidates besides
$0$ and $-1$ are the literals in the
clause. Any vote with $-1$ in first place
represents a truth assignment.
The positive literals in the truth
assignment are those Boolean variables
whose candidates appear
between $-1$ and $0$ in the vote.
With 2 candidates, $X$ and $Y$ both
elect the majority winner.
With 3 or more candidates,
$X$ elects
candidate $-1$ if there is a truth
assignment in the votes that satisfies
exactly one out of the three literals in
each clause represented by the votes
and otherwise elects $0$.
Computing a manipulation of $X$ 
is NP-hard.
Similarly, with 3 or more candidates,
$Y$ elects
candidate $0$ if there is a truth
assignment in the votes that satisfies
exactly one out of the 3 literals in
every clause represented by the votes
and otherwise elects $-1$.
Computing a manipulation of $Y$ 
is NP-hard. However,
$\winner{X}{Y}$ is polynomial to
manipulate since $0$ and $-1$ always
go through to the runoff where
the majority candidate wins.
\myqed

\begin{proposition}
There exist voting rules $X$ and $Y$
for which computing a manipulation
is polynomial but computing a manipulation
of $\winner{X}{Y}$ is NP-hard.
\end{proposition}
\myproof
The proof uses a similar reduction from 1 in 3-SAT on
positive clauses.
With 2 candidates, $X$ and $Y$ both
elect the majority winner.
With 3 or more candidates,
$X$ elects
candidate $-1$ if there is a truth
assignment in the votes that satisfies
at least one out of the three literals in
each clause represented by the votes
and otherwise elects $0$,
whilst $Y$ elects
candidate $-1$ if there is a truth
assignment in the votes that satisfies
at most one out of the three literals in
each clause represented by the votes
and otherwise elects $0$.
Computing a manipulation of $X$ or $Y$
is polynomial since we can simply construct
either the vote that 
sets each Boolean variable to true or to false.
However, computing a manipulation of $\winner{X}{Y}$
as it may require
solving a 1 in 3-SAT problem on positive
clauses.
\myqed

\subsection*{UNWEIGHTED VOTES, TRACTABLE CASES}
\vspace{-1mm}
If computing a manipulation of the base rules
is polynomial, it often remains polynomial to
compute a manipulation of the combined rules. However,
manipulations may now be more complex to compute.
We need to find a manipulation of one base rule
that is compatible with the other base rules, and
that also wins the runoff.
We illustrate this for various combinations of scoring rules.

\begin{proposition}
Computing a manipulation of $\winner{plurality}{veto}$
is polynomial.
\end{proposition}
\myproof\lirong{Added an overview of proof.}  We present a polynomial-time algorithm that checks whether $k$ manipulators can make $c$ win in the following two steps: we first check for every candidate $a$, whether the manipulators can make $c$ to be the plurality winner for $P\cup M$ while $a$ is the veto winner, and $c$ beats $a$ in the runoff (or $c=a$). Then, we check for every candidate $a$ whether the manipulators can make $c$ to be the veto winner while $a$ is the plurality winner, and $c$ beats $a$ in the runoff.

For the first step, let $S$ be a subset of candidates that beat $a$ in $P$
under $veto$.
We denote $\Delta_{s}$ as the difference
in the veto score of $s$, $s\in S$, and $a$ in $P$.
If the veto scores are equal and $s\succ a$
in the tie-breaking rule then we set $\Delta_{s} = 1$.
If $\sum_{s \in S} \Delta_{s} > k$ then $a$ can not win under $veto$.
Otherwise, we  place $s$ in
last positions in exactly $\Delta_{s}$ manipulator votes.
This placing is necessary for $a$ to win under $veto$.
We place $c$ in the first position and $a$ in the second position
in all votes in $M$. We fill the remaining positions arbitrarily.
This manipulation is optimal under an assumption
that $a$ wins under $veto$ as $c$ is always placed in the first position.
For each possible candidate for $a$, we check
if such a manipulation is possible and check if $c$ is the winner of
the run-off round. If we find a manipulation we stop.
The special case when $a=c$ is analogous.

For the second step, let $b$ be the candidate with maximum $plurality$ score in $P$.
We denote $\Delta_{a}$ to be the difference in the plurality scores
of $b$ and $a$.
We place $a$ in the $\Delta_{a}$ first positions in the manipulator votes.
The condition is necessary for $a$ to win under $plurality$.
We put $a$ in the second position in the remaining votes.
We put $c$ in the second position after $a$ in $\Delta_{a}$ manipulator
votes and put $c$ in the first position in the remaining $k-\Delta_{a}$
votes.
To ensure that  $c$ wins under $veto$ we perform the same procedure
as above. The only simplification is
that we do not need to worry about tie-breaking rule as
$c$ wins tie-breaking by assumption.
We fill the remaining positions arbitrarily.
This manipulation is optimal under an assumption
that $a$ wins under $plurality$, as $c$ is placed in the first position
unless $a$ has to occupy it.
For each possible candidate $a$ we check
if such a manipulation is possible and check if $c$ is the winner of
the run-off round. If we find a manipulation we stop. Otherwise,
there is no manipulation.
\myqed

It is also in P to decide if a single agent can manipulate
an election for any combination
of scoring rules. Interestingly, we can
use a perfect matching algorithm to compute
this manipulation.

\begin{proposition}
Computing a manipulation of $\winner{X}{Y}$ is
polynomial for a single manipulator and any pair of
scoring rules, $X$ and $Y$.
\end{proposition}
\myproof
Suppose there is a manipulating vote $v$ such that $c$
wins $P \cup \{v\}$ under $\winner{X}{Y}$. Let $X$ and $Y$
have the scoring vectors $(x_1,\ldots,x_m)$ and $(y_1,\ldots,y_m)$.
As is common in the literature, we assume tie-breaking
is in favour of $c$. Suppose $c$ wins under $X$
in a successful  manipulation. The case that $c$ wins
under $Y$ is dual.
Suppose another candidate $a$ wins
under $Y$, $c$ is placed at position $i$ and $a$ is placed at position
$j$ in $v$. We show how to construct this vote if it exists by
finding a perfect matching in a bipartite graph.
For each candidate besides $c$ and $a$, we introduce
a vertex in the first partition. 
For each position in $[1,m] \setminus \{i,j\}$
we introduce a vertex in the second partition. 
For each candidate $c_k$ besides $c$ and $a$
we connect the corresponding vertex with a vertex $t$ in the second
partition iff (1) the score of $c_k$ in $P$ under
$X$ less the
score of $c$ in $P$ under $X$ is less than
or equal to $x_i - x_k$,
and
(2) the score of $c_k$ in $P$ under
$Y$ less the score of $a$ in $P$ under $Y$ is less than
or equal to $y_j - y_t$,
or if two differences are equal then $a$ is before
$c_k$ in the tie-breaking rule.
In other words, we look for a placement of the
remaining candidates in
$v$ such that $c$ wins in $P\cup \{v\}$ under $X$,
$a$ wins in $P\cup \{v\}$ under $Y$, $c$ is at position $i$ and $a$
is at position $j$
in $v$. There exists a perfect matching in this graph iff
there is a manipulating vote
that satisfies our assumption.
If $a=c$, the reasoning is similar but we only need to fix
the position of $c$.
Using this procedure, we check for each candidate $a$
and for each pair of positions $(i,j)$
if there exists a vote $v$ such that
$c$ wins in $P\cup \{v\}$ under $X$,
$a$ wins in $P\cup \{v\}$ under $Y$, $c$ is at position $i$ and $a$ is at position $j$
in $v$. If such a vote exists,  we also check if $c$ beats $a$ in the run-off round.
If $c$ loses to $a$ in the run-off for all combinations of $a$ and $(i,j)$
then  no manipulation exists.
\myqed

\subsection*{UNWEIGHTED VOTES, INTRACTABLE CASES}
\vspace{-1mm}
We begin with combinations involving STV.
This was the first commonly used
voting rule shown to be NP-hard to manipulate
by a single manipulator \cite{stvhard}. Not surprisingly,
even when combined with voting rules which are polynomial
to manipulate like plurality, veto, or
$k$-approval, manipulation 
remains NP-hard
to compute.

\begin{proposition}
Computing a manipulation of $\winner{X}{STV}$
is NP-hard for $X \in \{plurality, \mbox{\it k-approval}, veto, Borda\}$
for $1$ manipulator.
\end{proposition}
\myproof (Sketch)
Consider the NP-hardness proof for manipulation
of STV~\cite{stvhard}.
We denote the profile constructed in the proof $P$.
The main idea is to modify $P$ so that the preferred
candidate $c$ can win under
$\winner{X}{STV}$ iff $c$ can win the modified election under $STV$.
For reasons of space, we illustrate this for $X=Borda$.
Other proofs are similar.
Candidate $w$ (who is the other possible winner of $P$)
has the top Borda score. Hence, $c$ must win
by winning the STV election (which is possible iff there
is a 3-cover). We still have the problem that
$w$ beats $c$ in the run-off. Hence, we
introduce
a dummy candidate $g'$ after $c$ in each
vote. This makes sure that the score of $g'$
is greater than or equal to $(n - 6)|P|$.
We also introduce $G = \big\lfloor\frac{|P|}{n}\big\rfloor$
blocks of $n$ votes.
Let
$P' = \bigcup_{k=1}^{G}  \bigcup_{i=1}^{n} (g'\succ  c_i \succ \ldots \succ c_{i-1}
)$.
The Borda score of $g'$ in $P \cup P'$ is greater
than that of any other candidate.
In an STV election on $P \cup P'$, $g'$ reaches
the last round.
Therefore, the elimination order remains
determined by the votes in $P$.
Hence, if there is a 3-cover, the candidate $c$ can reach
the last round.
In the worst case, when $|P|$ is divisible by $n$, the plurality
scores of $c$ and $g'$ are the same and $c$ wins 
by tie-breaking.
\myqed

We turn next to combinations of Borda voting, where it is NP-hard to manipulate
with two manipulators \cite{dknwaaai11,borda2}.

\begin{proposition}
Computing a manipulation of $\winner{X}{Borda}$ by two manipulators
is NP-hard for $X \in \{plurality, \mbox{\it k-approval}, veto\}$
\end{proposition}
\myproof (Sketch)
Consider the NP-hardness proof for manipulation of Borda
which uses a reduction from the permutation sums
problem \cite{dknwaaai11}.
Due to the spaces constraint, we consider only
$\winner{veto}{Borda}$. Other
proofs are similar but much longer and more complex.
The reduction uses
the following construction to
inflate scores to a desired target.
To increase the score of candidate $c_i$ by $1$ more
than candidates $c,c_1\ldots,c_{i-1},c_{i+1}, \ldots, c_{n-1}$ and by $2$ more
than candidate $c_{n}$ we consider the following pair of votes:
\vspace{-1mm}\begin{eqnarray*}
&c_i \succ c_{n} \succ c_1 \succ \ldots \succ c_{n-1} \\
&c_{n-1} \succ c_{n-2} \succ \ldots \succ c_1 \succ c_i \succ c_n
\end{eqnarray*}

\vspace{-1mm}We change the construction by putting $c$ in the last place
in the first vote in each pair of votes and first place in the second
vote, and leaving all other candidates unchanged when we increase
the score of $c_i \neq c$ by one.
This modification does not change the desired properties of these votes.
Note that $c$ and $c_n$ cannot be winners
under $veto$. Hence, $c$ must win under $Borda$ and then
win the run-off.
This is possible iff there exists a solution for permutation sums problem.
\myqed

\subsection*{WEIGHTED VOTES, TRACTABLE CASES}

We focus on elections with weighted votes
and 3 candidates.
This is the fewest number
of candidates which can give intractability.
All scoring rules besides
plurality (e.g. Borda, veto, 2-approval)
are NP-hard to manipulate in this case
\cite{dichotomy}.
We therefore focus on combinations of the voting rules:
plurality, cup, Copeland, maximin and Bucklin.
Computing a manipulation of each of these
rules is polynomial in this case.
We were unable to find a proof in the literature
that Bucklin is polynomial to manipulate with weighted votes,
so we provide one below.

\begin{proposition}
Computing a manipulation of $Bucklin$
is polynomial with weighted votes and 3 candidates.
\end{proposition}
\myproof
It is always optimal to place the
preferred candidate $c$ in the first
position as this only decreases the scores of
the other 2 candidates, $a$ and $b$.
We argue that the winner is
chosen in one of the first two rounds.
In the first round, if $c$ still loses to
$a$ or $b$ then there is no manipulation
that makes $c$ win.
In the second round, we must have at least
one candidate with a majority. Suppose we did not.
Then the sum of scores of the 3 candidates is at most
$3n/2$. But the sum of the approval votes is $2n$
which is a contradiction. Hence, if $c$ does not
get a majority in this round, one of the other
candidates wins regardless of the manipulating
votes.
%
\myqed

We recall that in this paper all ties are broken in favor of $c$, which is crucial in the proof of the above proposition. In fact, we can show that if some other tie-breaking mechanisms are used, then Bucklin is hard to manipulate with weighted votes, even for $3$ candidates. \lirong{I added this clarification.}

We next identify several cases where computing
a manipulation for combinations of
these voting rules is tractable.

\begin{proposition}
Computing a manipulation of $\winner{Copeland}{cup}$,
or of $\winner{Copeland}{Bucklin}$
is polynomial with
weighted votes and 3 candidates.
\end{proposition}
\myproof
First we consider the outcome of
$c$ vs $a$ and $c$ vs $b$
assuming that $c$ is ranked at the first position
by all manipulators.

{\bf Case 1.} Suppose $c$ is a Condorcet loser.
In this case, $c$ can only win
if $c$ wins under both $Copeland$ and $Y$.
However, $c$ must lose under $Copeland$ because
$Copeland$ never elects the Condorcet loser.

{\bf Case 2.}
Suppose $c$ is a Condorcet winner.
Then $c$ is a winner of both rules as they are
both Condorcet consistent.

{\bf Case 3.} Suppose there exists a candidate $a$ such that
$N_{P \cup M}(a,c) > N_{P \cup M}(c,a)$ and
$N_{P \cup M}(b,c) \leq N_{P \cup M}(c,b)$
even if $c$ is ranked first by all manipulators.
We argue that if there is a manipulation,
then all manipulators can vote $c \succ b \succ a$. We
consider the case that $c$ wins under $Copeland$
and $b$ wins under $cup$. The other cases ($b$ wins
under $Copeland$, $c$ wins under $Bucklin$, etc.) are
similar.
For $c$ to win under $Copeland$,
all candidates
have to have the $Copeland$ score of 0 as, by assumption,
$c$ loses to $a$. Hence, the maximum $Copeland$
score of $c$ is $0$. Therefore,
for $c$ to win  the following holds
$N_{E \cup M}(b,a) > N_{E \cup M}(a,b)$
and $N_{E \cup M}(c,b) > N_{E \cup M}(b,c)$.
The only possible agenda is $a$ v $c$, and the winner
playing $b$. In all other agendas, $b$ loses to $c$
in one of the rounds. For $b$ to win $cup$,
$N_{P \cup M}(b,a) \geq N_{P \cup M}(a,b)$ and tie-breaking
has $c \succ b \succ a$.
The manipulation vote $c \succ b \succ a$ will only help achieve
the inequalities in both cases.
\myqed

\begin{proposition}
Computing a manipulation of $\winner{Bucklin}{cup}$
is polynomial with weighted votes and 3 candidates.
\end{proposition}
\myproof
We consider three possible outcomes of pairwise
comparison between $c$ vs $a$ and $c$ vs $b$
assuming that $c$ is ranked at the first position
by all manipulators.

{\bf Case 1.} Suppose $c$ is a Condorcet loser after
the manipulation. $c$ can only win overall
if $c$ wins under both  $Bucklin$ and $cup$.
However, $c$ must lose under $cup$.

{\bf Case 2.}
Suppose $c$ is a Condorcet winner.
Then $c$ must be a winner of $cup$ as this is
Condorcet consistent. Hence, regardless
of the rest of the manipulating
votes, $c$ reaches the run-off round
and beats any other candidate.

{\bf Case 3.} Suppose there exists
candidate $a$ such that
$N_{P \cup M}(a,c) > N_{P \cup M}(c,a)$ and
$N_{P \cup M}(b,c) \leq N_{P \cup M}(c,b)$.
Note that $M$ must guarantee that
$a$ does not reach the run-off round as
$c$ loses to $a$ in the pairwise elections.
There are two sub-cases: $c$ wins under $cup$
and $b$ wins  under $Bucklin$ in $P\cup M$,
or $b$ wins under $cup$
and $c$ wins  under $Bucklin$.
As shown in the proof of the last Proposition,
if there is a manipulation, $c\succ b \succ a$ will work
in both cases.
\myqed

\vspace{-1mm}
\subsection*{WEIGHTED VOTES, INTRACTABLE CASES}
\vspace{-1mm}
We continue to focus on combinations of the voting rules:
plurality, cup, Copeland, maximin and Bucklin.
We give several results which show that
there exists combinations of these voting rules where
manipulation is intractable to compute 
despite the fact that all the base rules being
combined are polynomial to
manipulate. 
These results provide support
for our argument that combining voting rules is a
mechanism to increase the 
complexity
of manipulation.

\vspace{-3mm}
\begin{proposition}
Computing a manipulation of $\winner{plurality}{Y}$ where
$Y \in \{cup, Copeland, maximin, Bucklin\}$,
is NP-complete with weighted votes and 3 candidates.
\end{proposition}
\myproof (Sketch)
We consider the case $\winner{plurality}{cup}$. Other proofs are
similar but longer.
We reduce from a  {\sc  partition} problem in which we
want to decide if integers $k_i$ with sum
$2K$ divide into two equal sums of size $K$.
Consider the following profile:

\vspace{-5mm}\begin{align*}
4K&a \succ b \succ c \hspace{10mm}
4K&a \succ c \succ b \\
1K&b \succ a \succ c \hspace{10mm}
9K&b \succ c \succ a
\end{align*}

\vspace{-2mm}For each
integer $k_i$, we have a member of the manipulating coalition
with weight $2k_i$.
The tie-breaking rule is $c \succ a \succ b$.
The $cup$ has $a$ play $b$, and the winner
meets $c$. Note that $b$ cannot reach the
run-off as they beat $c$ in pairwise
elections whatever the manipulators do.
Note that $c$ cannot win the $plurality$
rule. Hence $a$ must be the $plurality$ winner.
The run-off is $a$, the $plurality$ winner
against $c$, the $cup$ winner (which is the
same as the final round of the $cup$).
For this to occur, the manipulators have to partition
their votes so that exactly $2K$ manipulators put $c$
above $a$ and $2K$ 
put $a$
in the first position (and above $c$).\footnote{Here we abuse the notation by saying ``$2K$ manipulators'', which we meant ``manipulators whose weights sum up to $2K$''.} Therefore there exists a manipulation
iff there exists a partition.
\myqed.

\vspace{-3mm}
\begin{proposition}
Computing a manipulation of $\winner{Copeland}{Y}$ where
$Y \in \{plurality, maximin\}$,
is NP-complete with weighted votes and 3 candidates.
\end{proposition}
\myproof (Sketch)
We consider the case $\winner{Copeland}{plurality}$.
Other proofs are similar but longer.
We again reduce from a  {\sc  partition} problem.
Consider the following profile:
\vspace{-2mm}\begin{eqnarray*}
7K&b \succ c \succ a \hspace{10mm}
K&b \succ a \succ c \\
4K&a \succ c \succ b \hspace{10mm}
2K&a \succ b \succ c \hspace{10mm}
1\ \ \  c \succ a \succ b
\end{eqnarray*}

\vspace{-2mm}For each
integer $k_i$, we have a member of the manipulating coalition
with weight $2k_i$.
Now, $b$ must not reach the run-off round and
$a$ must win $plurality$ by similar arguments
to the last proof. Hence $c$ must be the $Copeland$
winner.
For this to occur, the manipulators have to partition
their votes so that exactly $2K$ manipulators put $c$
above $a$, $2K$ manipulators put $a$
in the first position (and above $c$)
and put $b$ in the last position in all votes.
 Therefore there exists a manipulation
iff there exists a partition.
\myqed
\vspace{-2mm}
\begin{proposition}
Computing a manipulation of $\winner{maximin}{Y}$ where
$Y \in \{plurality, cup, Copeland, Bucklin\}$,
is NP-complete with weighted votes and 3 candidates.
\end{proposition}
\myproof (Sketch)
We consider the case $\winner{maximin}{plurality}$. Other
proofs are similar but longer.
We reduce from a  {\sc  partition} problem in which we
want to decide if integers $k_i$ with sum
$2K$ divide into two equal sums of size $K$.
Consider the following profile:
\vspace{-5mm}\begin{eqnarray*}
4K&b \succ c \succ a \hspace{10mm}
2K&b \succ c \succ a \\
2K&a \succ b \succ c \hspace{10mm}
2K&a \succ c \succ b
\end{eqnarray*}

\vspace{-3mm}For each
integer $k_i$, we have a member of the manipulating coalition
with weight $2k_i$.
Now, $b$ must not reach the run-off round and
$a$ must win $plurality$ by similar arguments
to the last proof. Hence $c$ must be the $maximin$
winner.
For $a$ to win $plurality$, manipulators
with total weight at least $2K$ must rank $a$ first.
Before the manipulators vote,
the $maximin$ score of $a$ is $4K$,
of $b$ is $6K$ and of $c$ is $2K$.
We note that $c$ must be ranked above $b$ in all manipulators
votes and above $a$ in $2K$ manipulators votes,
otherwise $c$ loses to $b$ under $maximin$.
As $2K$ manipulators
must vote $a \succ\ c \succ b$,
we have $N_{P \cup M}(a,b) \geq 6K$,
$N_{P \cup M}(c,b) \geq 4K$ and $N_{P \cup M}(a,c) \geq 6K$.
This increases the $maximin$ score of $a$ to $6K$
and of $c$ to $4K$.
Now $c$ must be ranked
above $a$ in at least $2K$ manipulators votes to increase
its $maximin$ score to $6K$. Hence, the only possible
option is if $2K$ manipulators
vote $a \succ c \succ b$ and $2K$ vote
and $c \succ  a \succ b$ with weight $2K$.
In this case the $maximin$ score of all candidates are the same
and equal to $6K$. By the tie-breaking rule, $c$ wins.
Therefore, there exists a manipulation iff there exists a partition.
\myqed

We summarize our results about weighted manipulation in
the following table.

\begin{table}
{
  \centering
  \begin{tabular}{|c|c|c|c|c|c|c|c|}
    \hline
$\winner{X}{Y}$ & plurality        &maximin      & Copeland       & cup                & Bucklin \\
      \hline
plurality &P &NPC
                     &NPC
                                                       &  NPC&  NPC\\
      \hline
      maximin &--& P & NPC & NPC    &  NPC       \\
      \hline
     Copeland &--& --& P & P   &      P       \\
      \hline
      cup &--&--&--                           &    P  &  P       \\
      \hline
      Bucklin &--&--&--                          &                       --          &  P    \\
              \hline
  \end{tabular}
  }
  \caption{Computational complexity of coalition
manipulation with weighted votes and 3 candidates}\label{table:weighted}
\end{table}

\begin{table*}[htp]
  \centering
  \begin{tabular}{cc}
\begin{tabular}{|c|c|c|c|c|c|}
\hline
$n,m$ & Opt & $\bordashort$ & $\plurshort$ & $\bordaplurshort$   \\
\hline
\hline
  4,4& \textbf{1.17 } & 1.44  & 1.81  & 1.41  \\
  4,8& \textbf{1.78 } & 2.02  & 2.52  & 2.02  \\
  4,16& \textbf{1.94 } & 2.25  & 2.63  & 2.19  \\
  4,32& \textbf{2.85 } & 3.17  & 3.99  & 3.07  \\
  8,8& -- & 2.00  & 2.83  & \textbf{1.81 } \\
  8,16& -- & 2.44  & 3.36  & \textbf{2.21 } \\
  8,32& -- & 3.63  & 5.25  & \textbf{3.13 } \\
  16,16& -- & 2.85  & 4.32  & \textbf{2.27 } \\
  16,32& -- & 3.71  & 6.11  & \textbf{2.92 } \\
  32,32& -- & 3.96  & 6.36  & \textbf{2.88 } \\
\hline
\end{tabular} & \begin{tabular}{|c|c|c|c|c|c|}
\hline
$n,m$ & Opt & $\bordashort$ & $\plurshort$ & $\bordaplurshort$   \\
\hline
\hline
  4,4& \textbf{2.19 } & 2.35  & 2.76  & 2.34  \\
  4,8& \textbf{3.79 } & 4.09  & 4.50  & 4.03  \\
  4,16& \textbf{7.68 } & 7.99  & 8.59  & 7.92  \\
  4,32& \textbf{11.68 } & 12.20  & 13.07  & 11.94  \\
  8,8& -- & 4.42  & 5.25  & \textbf{4.21 } \\
  8,16& -- & 8.41  & 10.06  & \textbf{8.12 } \\
  8,32& -- & 17.79  & 20.85  & \textbf{16.95 } \\
  16,16& -- & 9.73  & 11.83  & \textbf{9.18 } \\
  16,32& -- & 16.98  & 21.07  & \textbf{15.79 } \\
  32,32& -- & 18.66  & 23.28  & \textbf{17.64 } \\
\hline
\end{tabular}\\ (a) The uniform model. & (b) The urn  model.
\end{tabular}
\caption{\label{t:t1} Experiments on randomly generated profiles: average
number of required manipulators.} 
\end{table*}

\section{APPROXIMATION}

One way to deal with the
intractability of manipulation is to view
computing a manipulation as an approximation problem where we try to
minimise the number of manipulators. 
We argue here that combining
voting rules together can make
such approximation problems more
challenging. In particular, we show that
a good approximation method for
a rule like $Borda$ will perform
very poorly when $Borda$ is combined with
a simple rule like $plurality$ or $veto$.
We consider the $Greedy$ algorithm for
Borda that computes a manipulation that
is within 1 of the optimal
number of manipulators \cite{zuckermanSODA08}. 

\begin{proposition}
There exists a family of profiles such that the
$Greedy$ approximation method
on $\winner{X}{Borda}$ requires
$k + \Omega(|P|)$ manipulators where $k$ is an
optimum number of manipulators for $Borda$, $P$ is
the profile in question and $X = \{plurality, veto\}$
\end{proposition}
\myproof
We consider $\winner{veto}{Borda}$. A similar
argument holds for $\winner{plurality}{Borda}$.
Consider the following profile:
$(c_{i\mytext{ mod } n} \succ c_{(i+1)\mytext{ mod } n}\succ \ldots \succ c_{n+i-1 \mytext{ mod } n})$ for $i=0,\ldots,n-1$,
and 1 vote
for $(c_{n-1}\succ \ldots  \succ c_{1} \succ c_0)$.
The tie-breaking rule is $c_0 \succ c_{n-1} \succ \ldots \succ
c_{1}$ where
the preferred candidate is $c_0$.
The score of the candidates $c_i$ is $n(n-1)/2 + i$, $i=0,\ldots,n-1$.
The $Greedy$ algorithm outputs the vote
$(c_0 \succ c_1 \succ \ldots  \succ  c_{n-1})$.
This increase the number of veto points of $c_{n-1}$ by 1.
Note that $c_i$, $i=1,\ldots,n-2$ all have only one veto point.
Hence, $c_{n-2}$ wins by the tie breaking.
Note that $c_0$ has 2 veto-points.
However, $c_0$ loses to $c_{n-2}$ in the run-off round
as $c_{n-2}$ is ranked before $c_0$ in $n-2$ votes.
Hence, the $Greedy$ algorithm will continue to produce pairs,
$(c_0 \succ c_1\succ \ldots  \succ c_{n-1})$ and
$(c_0 \succ c_{n-1}\succ \ldots  \succ c_{1})$,
until $c_0$ takes first positions in $n-1$ votes and wins the run-off round.
On the other hand, if we add the votes
$(c_0 \succ c_1 \succ \ldots \succ  c_{i-1} \succ c_{i+1} \succ \ldots \succ  c_{n-1} \succ c_i)$ for $i=\lceil n/2 \rceil,\ldots, n-2$
then we increase veto points of candidates $c_{\lceil n/2 \rceil}, \ldots, c_{n-2}$ by one.
Hence, $c_{\lfloor n/2 \rfloor}$ wins under $veto$
by tie-breaking, and then loses to  $c_0$  in
the run-off.
\myqed

\vspace{-2mm}
\section{EXPERIMENTAL RESULTS}
\vspace{-1mm}

We investigated the effectiveness of approximation
methods on combinations of voting rules experimentally.
We used a similar setup
to~\cite{dknwaaai11}.  We generated
uniform random votes and votes drawn from the Polya
Eggenberger urn model. 
In the urn model,
votes are drawn from an urn at random, and are placed back into
the urn along with $a$ other votes of the same type.  This captures
varying degrees of social homogeneity. We set $a = m!$
so that there is a 50\% chance that the second vote is the same as the
first. For each combination of the number of candidates $n$, $n \in \{4,8,16,32\}$,
and the number of voters, $m$, $ m \in \{4,8,16,32\}$
and $n \leq m$, we generated
200 instances of elections for each model.

We ran four algorithms. The first algorithm,\emph{ Opt},
finds an optimum solution of the manipulation problem for
$\winner{plurality}{Borda}$.
We use a constraint solver to encode the
manipulation problem as a CSP. We could only solve small
problems using complete search as the CSP model is
loose. 
The second algorithm is $\bordashort$ algorithm from~\cite{zuckermanSODA08}
that we run until the winner
it produces is also a winner of $\winner{plurality}{Borda}$. The third
algorithm, $\plurshort$, is a greedy algorithm for $plurality$.
Again, we run until its output is a winner
of $\winner{plurality}{Borda}$. The fourth algorithm, $\bordaplurshort$,
 is our modification
of $\bordashort$ that simultaneously tries to manipulate
$Borda$ and $plurality$. The algorithm runs the $\bordashort$
heuristic first and checks if the preferred candidate $c$ is
a winner under $\winner{plurality}{Borda}$. If $c$ loses under both
rules we increase the number of manipulators. If $c$
wins under one of the rules we check if we can make
a candidate $a \in \mathcal{C} \setminus c$ the winner
of the other rule, where $\mathcal{C}$ is the set of candidates.
If we want $a$ to win under $plurality$
then we place $a$ in exactly as many first positions
as it needs to win under $plurality$ and place $c$ in the remaining
first and second positions. We run $\bordashort$
to place the remaining  positions starting with
votes where the first two positions are fixed.
If we want $a$ to win  under $Borda$, we
find the maximum number of first
positions for $c$ such that $a$ still can win
under $Borda$ and fill the remaining positions
using $\bordashort$. In both case, we check that
the preferred candidate is winner under
$\winner{plurality}{Borda}$.  If not, we increase the number
of manipulators. Tables~\ref{t:t1}
show the results of our experiments.
First, they show that we need to adapt
approximation algorithms
for individual rules to obtain a
solution that is close to the optimum number
of manipulators. As the size
of the election grows individual
approximation algorithms require
significantly more manipulators
than the optimum.
Second, for the combination of $plurality$
and $Borda$, our adaptation
of the $\bordashort$ method works very well
and finds a good approximation.
Experimental results suggest
that 
it finds a solution with at most one additional
manipulator.

\vspace{-1mm}
\section{CONCLUSION}

We have put forwards a simple method for combining together
voting rules that performs a run-off between
the different winners of each voting rule.
We have provided theoretical and experimental
evidence for the value of this combinator. On the
theoretical side, we proved that a combination
of voting rules can inherit a desirable property like
Condorcet consistency or the majority criterion
from just one base voting rule. On the other
hand, two important properties can be lost by the
introduction of a run-off: monotonicity and
consistency. Combining voting rules also
tends to increase the computational difficulty
of finding a manipulation. For instance,
with weighted votes, we proved that
computing a manipulation for a simple
combination like $plurality$ and $cup$
is NP-hard, even though
$plurality$ and $cup$ on their own
are polynomial to manipulate.
On the experimental side,
we studied the impact of this combinator
on approximation methods that
find close to optimal manipulations.



\vspace{-1mm}
\begin{thebibliography}{10}

\bibitem{stvhard}
J.J. Bartholdi and J.B. Orlin, `Single transferable vote resists strategic
  voting', {\em Social Choice and Welfare}, {\bf 8}(4),  341--354, (1991).

\bibitem{bartholditoveytrick}
J.J. Bartholdi, C.A. Tovey, and M.A. Trick, `The computational difficulty of
  manipulating an election', {\em Social Choice and Welfare}, {\bf 6}(3),
  227--241, (1989).

\bibitem{borda2}
N.~Betzler, R.~Niedermeier, and G.J. Woeginger, `Unweighted coalitional
  manipulation under the {Borda} rule is {NP}-hard', in {\em Proc.~of
IJCAI}, pp. 55--60,
  (2011).

\bibitem{csijcai03}
V.~Conitzer and T.~Sandholm, `Universal voting protocol tweaks to make
  manipulation hard', in {\em Proc.~of  IJCAI}, pp. 781--788,
 (2003).

\bibitem{csljacm07}
V.~Conitzer, T.~Sandholm, and J.~Lang, `When are elections with few candidates
  hard to manipulate', {\em JACM}, {\bf 54} (3), 1--33, (2007).

\bibitem{dknwaaai11}
J.~Davies, G.~Katsirelos, N.~Narodytska, and T.~Walsh, `Complexity of and
  algorithms for {Borda} manipulation', in {\em Proc.~of AAAI}, pp. 657--662, (2011).

\bibitem{elisaac05}
E.~Elkind and H.~Lipmaa, `Hybrid voting protocols and hardness of
  manipulation', in {\em Proc.~of ISAAC'05}, pp. 24--26,(2005).

\bibitem{dichotomy}
E.~Hemaspaandra and L.A. Hemaspaandra, `Dichotomy for voting systems', {\em
  Journal of Computer and System Sciences}, {\bf 73}(1),  73--83, (2007).

\bibitem{wisdomcrowds}
James Surowiecki, {\em The Wisdom of Crowds: Why the Many Are Smarter Than the
  Few and How Collective Wisdom Shapes Business, Economies, Societies and
  Nations}, Little Brown \& Co, 2004.

\bibitem{wxaamas12}
T.~Walsh and L.~Xia, `Lot-based voting rules', in {\em Proc.~ of AAMAS},
 (2012).

\bibitem{zuckermanSODA08}
M.~Zuckerman, A.D. Procaccia, and J.S. Rosenschein, `Algorithms for the
  coalitional manipulation problem', in {\em Proc.~of SODA}, pp. 277--286, (2008).

\end{thebibliography}

\end{document}